\documentclass[runningheads,a4paper,english]{llncs}[2022/01/12]

\usepackage{upquote}
\usepackage{amsmath,amssymb,amsfonts}
\usepackage{graphicx}
\usepackage{textcomp}
\usepackage[utf8]{inputenc}
\usepackage{xcolor}
\usepackage{adjustbox} 
\usepackage{makecell} 

\usepackage[ngerman,main=english]{babel}
\usepackage[ruled]{algorithm2e} 
\usepackage{algorithmic}
\usepackage{multirow}
\usepackage{caption}
\usepackage{subfloat}
\usepackage{pgfplots}
\addto\extrasenglish{\languageshorthands{ngerman}\useshorthands{"}}
\usepackage{regexpatch}
\makeatletter
\edef\switcht@albion{%
  \relax\unexpanded\expandafter{\switcht@albion}%
}
\xpatchcmd*{\switcht@albion}{ \def}{\def}{}{}
\xpatchcmd{\switcht@albion}{\relax}{}{}{}
\edef\switcht@deutsch{%
  \relax\unexpanded\expandafter{\switcht@deutsch}%
}
\xpatchcmd*{\switcht@deutsch}{ \def}{\def}{}{}
\xpatchcmd{\switcht@deutsch}{\relax}{}{}{}
\edef\switcht@francais{%
  \relax\unexpanded\expandafter{\switcht@francais}%
}
\xpatchcmd*{\switcht@francais}{ \def}{\def}{}{}
\xpatchcmd{\switcht@francais}{\relax}{}{}{}
\makeatother

\usepackage[hyphens]{url}

\makeatletter
\g@addto@macro{\UrlBreaks}{\UrlOrds}
\makeatother

\usepackage[%
    rm={oldstyle=false,proportional=true},%
    sf={oldstyle=false,proportional=true},%
    tt={oldstyle=false,proportional=true,variable=false},%
    qt=false%
]{cfr-lm}

\usepackage[T1]{fontenc}

\usepackage[
  babel=true, 
  expansion=alltext,
  protrusion=alltext-nott, 
  final 
]{microtype}

\DisableLigatures{encoding = T1, family = tt* }

\usepackage{graphicx}

\usepackage{diagbox}

\usepackage{xcolor}

\usepackage{listings}

\definecolor{eclipseStrings}{RGB}{42,0.0,255}
\definecolor{eclipseKeywords}{RGB}{127,0,85}
\colorlet{numb}{magenta!60!black}

\lstdefinelanguage{json}{
    basicstyle=\normalfont\ttfamily,
    commentstyle=\color{eclipseStrings}, 
    stringstyle=\color{eclipseKeywords}, 
    numbers=left,
    numberstyle=\scriptsize,
    stepnumber=1,
    numbersep=8pt,
    showstringspaces=false,
    breaklines=true,
    frame=lines,
    string=[s]{"}{"},
    comment=[l]{:\ "},
    morecomment=[l]{:"},
    literate=
        *{0}{{{\color{numb}0}}}{1}
         {1}{{{\color{numb}1}}}{1}
         {2}{{{\color{numb}2}}}{1}
         {3}{{{\color{numb}3}}}{1}
         {4}{{{\color{numb}4}}}{1}
         {5}{{{\color{numb}5}}}{1}
         {6}{{{\color{numb}6}}}{1}
         {7}{{{\color{numb}7}}}{1}
         {8}{{{\color{numb}8}}}{1}
         {9}{{{\color{numb}9}}}{1}
}

\usepackage[autostyle=true]{csquotes}

\defineshorthand{"`}{\openautoquote}
\defineshorthand{"'}{\closeautoquote}

\usepackage{booktabs}

\usepackage{paralist}

\usepackage[%
  square,        
  comma,         
  numbers,       
  sort&compress 
]{natbib}

\usepackage{etoolbox}
\makeatletter
\patchcmd{\NAT@test}{\else \NAT@nm}{\else \NAT@hyper@{\NAT@nm}}{}{}
\makeatother

\SetExpansion
[ context = sloppy,
  stretch = 30,
  shrink = 60,
  step = 5 ]
{ encoding = {OT1,T1,TS1} }
{ }

\usepackage[rflt]{floatflt}

\usepackage{pdfcomment}

\usepackage{stfloats}
\fnbelowfloat

\usepackage[group-minimum-digits=4,per-mode=fraction]{siunitx}
\addto\extrasgerman{\sisetup{locale = DE}}

\usepackage{hyperref}

\hypersetup{
  hidelinks,
  colorlinks=true,
  allcolors=black,
  pdfstartview=Fit,
  breaklinks=true
}

\usepackage[all]{hypcap}

\usepackage[caption=false,font=footnotesize]{subfig}

\usepackage[capitalise,nameinlink]{cleveref}

\crefname{section}{Sect.}{Sect.}
\Crefname{section}{Section}{Sections}
\crefname{listing}{List.}{List.}
\crefname{listing}{Listing}{Listings}
\Crefname{listing}{Listing}{Listings}
\crefname{lstlisting}{Listing}{Listings}
\Crefname{lstlisting}{Listing}{Listings}

\usepackage{lipsum}

\usepackage[math]{blindtext}
\usepackage{mwe}
\usepackage[realmainfile]{currfile}
\usepackage{tcolorbox}
\tcbuselibrary{listings}

\DeclareFontFamily{U}{MnSymbolC}{}
\DeclareSymbolFont{MnSyC}{U}{MnSymbolC}{m}{n}
\DeclareFontShape{U}{MnSymbolC}{m}{n}{
  <-6>    MnSymbolC5
  <6-7>   MnSymbolC6
  <7-8>   MnSymbolC7
  <8-9>   MnSymbolC8
  <9-10>  MnSymbolC9
  <10-12> MnSymbolC10
  <12->   MnSymbolC12%
}{}
\DeclareMathSymbol{\powerset}{\mathord}{MnSyC}{180}

\usepackage{xspace}

\makeatletter
\newcommand{\hydash}{\penalty\@M-\hskip\z@skip}
\makeatother

\input glyphtounicode
\begin{document}

\title{Towards Robust Recommender Systems via Triple Cooperative Defense}

\author{Qingyang Wang\inst{1} \and
	Defu Lian\thanks{Corresponding author}\inst{1} \and
	Chenwang Wu\inst{1} \and
	Enhong Chen\inst{1}}
%
%
\institute{University Of Science And Technology Of China, 96 Jinzhai Road, Hefei, Anhui, China\\
\email{greensun@mail.ustc.edu.cn, liandefu@ustc.edu.cn, wcw1996@mail.ustc.edu.cn, cheneh@ustc.edu.cn}}

\maketitle

\pagestyle{empty}  
\thispagestyle{empty} 

\begin{abstract}
Recommender systems are often susceptible to well-crafted fake profiles, leading to biased recommendations. The wide application of recommender systems makes studying the defense against attack necessary. Among existing defense methods, data-processing-based methods inevitably exclude normal samples, while model-based methods struggle to enjoy both generalization and robustness. Considering
the above limitations, we suggest integrating data processing and robust model and propose a general framework, Triple Cooperative Defense (TCD), which cooperates to improve model robustness through the co-training of three models. Specifically, in each round of training, we sequentially use the high-confidence prediction ratings (consistent ratings) of any two models as auxiliary training data for the remaining model, and the three models cooperatively improve recommendation robustness. Notably, TCD adds pseudo label data instead of deleting abnormal data, which avoids the cleaning of normal data, and the cooperative training of the three models is also beneficial to model generalization. Through 
extensive experiments with five poisoning attacks on three real-world datasets, the results show that the robustness improvement
of TCD significantly outperforms baselines. It is worth mentioning that TCD is also beneficial for model generalizations.
\end{abstract}


\section{Introduction}
In recent years, with the rapid development of Internet technology, the amount of information on the Internet has shown explosive growth. To obtain valuable information from massive data information more quickly and effectively, ``recommender systems''\cite{bobadilla2013recommender} came into being and quickly gained extensive attention and practical application in academia and industry. Recommender algorithms mine the content that the user is interested in from a large amount of data by using information such as user behavior and item characteristics and presenting it to the user in a list\cite{himeur2022blockchain}. Their superiority and commercial background make them widely used in various industries \cite{lian2020geography, bobadilla2013recommender,chevalier2006effect}.

However, the recommender system also faces the test of severe security problems while providing convenience for our lives. Since the collaborative filtering method works based on user profile information, it is easily affected by false user profile information. Studies \cite{wu2021triple,li2016data,lin2020attacking} have long shown that
recommender systems, especially those in the field of sales
and scoring, systematically interfere with the user ratings
included in the system, which will also impact users’ purchase
behavior and system recommendation results \cite{chevalier2006effect}. And even if attackers do not know the algorithm or implementation details used by the recommendation system, only using small-scale misleading data, can also have obvious interference effects on the normal recommendation behavior of the system, (e.g., in 2002,  after receiving a complaint, Amazon found that when a website recommends a Christian classic, another irrelevant book will be recommended simultaneously, which is caused by malicious users using deceptive means \cite{liu2014new}).

Two main defense methods against poisoning attacks are data-processing-based defense and model-based defense \cite{wu2021fight,deldjoo2021survey}. Data-based defense tries to study the characteristics of poisoning attacks, strip fake profiles, and purify datasets before the training of recommender systems. However, to pursue high recall, these methods will inevitably delete normal data, which will lead to biased recommendations. Model-based defense improves the robustness of the recommendation algorithm itself, and adversarial training \cite{madry2017towards} is recognized as the most popular and effective model-based defense method to enhance recommendation robustness \cite{wu2021fight}. This method maximizes recommendation error while minimizing the model's empirical risk by adding adversarial perturbations to the model parameters, eventually building robust models in adversarial games. Although adversarial training can significantly improve the robustness of the recommender system, it is difficult to control the strength of adversarial noise, which results in reducing the generalization of the recommendation to a certain extent. Besides, a recent study has shown that adversarial training with perturbations added to model parameters cannot well resist poisoning attacks \cite{wu2021fight}. Therefore, it is very needed to design a suitable means to integrate them and make use of their strengths and avoid weaknesses.

Based on the shortcomings mentioned above, we propose a novel defense method that integrates data processing and model robustness boosting, Triple Cooperative Defense(TCD), to enhance the robustness of recommender systems. Specifically, in each round of training, we sequentially use the high-confidence prediction ratings (consistent ratings) of any two models as auxiliary training data for the remaining models, and the three models cooperatively improve recommendation robustness. The proposed strategy is based on the following considerations. In the recommender system,  extremely sparse user-item interactions are difficult to support good model training, leading to models that are easily misled by malicious profiles. Besides, recent work also emphasizes that the model's robustness requires more real data\cite{wu2021fight}.  Therefore, we make reasonable use of cheap pseudo-labels. Obviously, pseudo-labels must be guaranteed by high-confidence ratings, but in the explicit feedback-based recommender system that we focus on, the predicted value is the rating, not the confidence. To this end, we suggest training with three models and any two models' consistent prediction ratings as auxiliary training data for the third model. Model robustness is improved in data augmentation and co-training of the three models. Notably, we do not cull the data nor modify the individual model structure, which can overcome the shortcomings of existing defense methods. Through 
extensive experiments with five poisoning attacks on three real-world datasets, the results show that the robustness improvement
of TCD significantly outperforms baselines. It is worth mentioning that TCD also improves model generalization.

The main contributions of this work are summarized as follows:
\begin{itemize}
	\item the proposal of a novel robust training strategy, named
	Triple Cooperative Defense, by generating pseudo labels into the recommender system for eliminating the damage of malicious profiles to models, and training three models cooperatively for improving model robustness. It is noteworthy that this is the first algorithm to combine data-processing-based defense and model-based defense in recommender systems. 
	\item an extensive study of co-training (defensive) methods to robustify the recommendation performance through the analysis of five attacks and three recommendation datasets. The results verify that our method enhances the robustness of the recommendation while ensuring generalization.
\end{itemize}

\section{Related Work}
\subsection{Security of Recommender Systems}
Many issues about security and privacy have been studied in recommender systems, which suggest that  recommender systems are vulnerable \cite{du2018enhancing, si2020shilling}, which leads to developing a toolkit for evaluating robustness \cite{ovaisi2022rgrecsys}.
Earlier attacks injected malicious profiles manually generated with little knowledge about the recommender system,
so it could not achieve satisfactory attack performance, e.g., random
attack\cite{lam2004shilling} and average attack \cite{lam2004shilling}.
The training of model-based recommendation algorithms usually
used backpropagation \cite{guo2017deepfm, he2017neural}, so perturbations were added along
the gradient direction to perform the attack \cite{fang2020influence, fang2018poisoning,li2016data,tang2020revisiting}. Inspired
by the GAN’s application \cite{jin2020sampling} in the recommendation, some work\cite{christakopoulou2019adversarial, lin2020attacking} used GAN to generate real-like fake ratings to bypass the
detection. With
the development of optimization algorithms, many works focused on attack specific types of recommender systems and turned attacks into optimization problems of deciding appropriate rating scores for users \cite{lam2004shilling,li2016data,yang2017fake,fang2018poisoning, oh2022robustness}. Moreover, some works \cite{fan2021attacking,song2020poisonrec}treated the items’ ratings as actions and used reinforcement learning to generate real-like fake ratings. Such optimization-based methods have strong attack performance, so defense is needed to mitigate the harm of attack.

\subsection{Defense against Poisoning Attacks}
According to the defense objective, a defense can be (i) reactive attack detection\cite{deldjoo2021survey} or (ii) proactive robust model construction, which will be listed below.

Many researchers used KNN, C4.5, and SVM \cite{burke2006classification}to supervise the statistical attributes to detect attacks. In most practical recommendation systems, due to the small number of labeled users and the lack of prior knowledge, unsupervised learning \cite{zhang2018ud,zhang2014detection}and
semi-supervised learning \cite{cao2013shilling} were used to detect attacks. However, to pursue high recall, these methods inevitably delete normal data, which lead to biased recommendations. Conversely, for our proposed TCD to enrich high-confidence data rather than remove outliers, it can avoid cleaning normal data and train a more accurate and robust model.

Athalye et al.\cite{athalye2018obfuscated} proposed defenses based on gradient masking produce models containing smoother gradients that hinder
optimization-based attack algorithms from finding the wrong directions in space\cite{machado2021adversarial}. More recently,
many works\cite{du2018enhancing,he2018adversarial,li2020adversarial,park2019adversarial,tang2019adversarial} have focused on adversarial
training. Assuming that each instance may be the target of attacks
\cite{machado2021adversarial}, adversarial training adds perturbations to the inputs or model
parameters that force the model to learn fragile perturbations. Although adversarial training can significantly improve the robustness of recommender systems, it is difficult to control the strength of adversarial data, which results in reducing the generalization of the recommendation. Instead, the proposed TCD does not need to add sensitive noise and is trained cooperatively to facilitate generalization, and we will prove it in section 4.

\section{Methodology}

\subsection{Threat Model}
\subsubsection{Attack goal}
Different shilling attacks may have different intents, but the eventual goal of an attacker may be one of several alternatives. We can divide the attack intents into three types, including push attacks, nude attacks, and random vandalism \cite{si2020shilling}. The push attack (nude attack) typically aims to increase (decrease) the popularity of the target item. For the random vandalism, the attacker combines push attack and null attack to maximize the error of the recommendation making users stop trusting the recommendation model and finally stop using it. We mainly focus on the defense against push attack, while nude attacks can be achieved by increasing the popularity
of non-target items until the target item is not in the user’s recommendation list \cite{yang2017fake}, which in a sense is equivalent to push
attacks. 
\subsubsection{Attack knowledge-cost}
Attacker’s knowledge-cost can be divided into high-knowledge attacks and low-knowledge
attacks\cite{si2020shilling}. The former requires the attackers to know detailed knowledge of the
rating distribution in a recommender system’s database, such as the algorithm used, specific parameter
settings, and even the users’ historical behavior, the latter only knows system-independent knowledge such as knowledge
might be obtained by consulting public information sources. Obviously, low-knowledge attacks are more practical because it is difficult for attackers to obtain detailed data and models. Therefore, we study the robust defense against low-knowledge attacks.
\subsubsection{Attack size}
Attack size is the number of fake profiles injected into the system by the attackers \cite{wu2014survey}. Obviously, the model
robustness and attack size cannot be decoupled. Considering that most users only rate a small number of items, the greater the attack intensity, the more likely it is to be detected \cite{mobasher2007toward}. Similar to \cite{wu2021fight}, we limit the attacker size to 5\%, and the limit of the number of ratings for each attacker is the average number of ratings.

\subsection{Triple Cooperative Defense}
\begin{figure}[h]
	\centering
	\includegraphics[width=1\columnwidth]{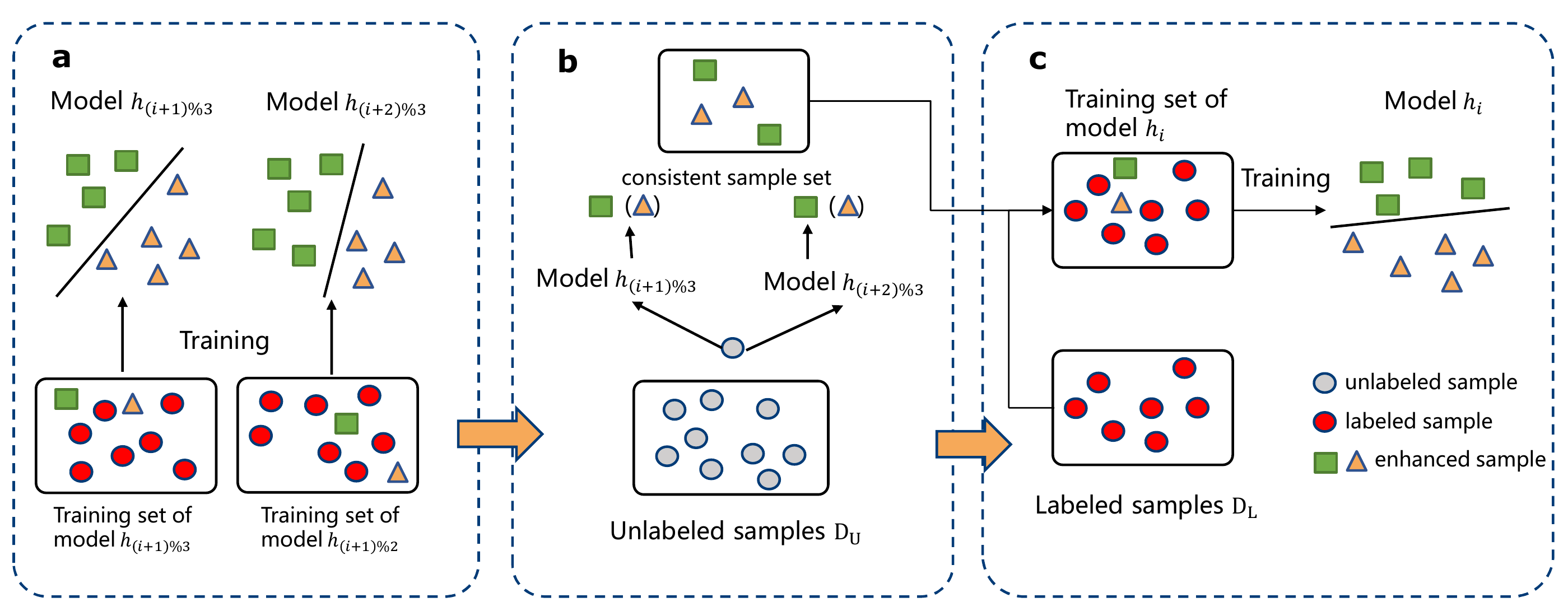}
	\captionsetup{font=scriptsize}
	\caption{The training of model $h_{i}$ in each round. a: The other two models use the same collaborative training. b: The labels predicted the same by the two models are taken as consistent samples. c: Model i is trained on labeled samples $D_{L}$ and consistent samples.}
	\label{fig:framework}
\end{figure}

As discussed in section 1, data-processing-based defense inevitably removes normal data altogether to achieve high recall rates, while model-based defense is difficult to enjoy both robustness and generalization\cite{zhang2020attacks}. Therefore, it is crucial to combine them effectively and design a defense algorithm that maximizes their strengths and circumvents their weaknesses. Recent studies\cite{deldjoo2021survey} demonstrated that robust models require more labeled data\cite{wu2021fight}. Besides, the recommender system is extremely sparse, that is, there is little interactive information about users and items, making a small amount of normal data difficult to support good training of the model, and maybe misled easily by malicious data and produce biased recommendations. This finding makes us reasonably believe that the vulnerability of the recommendation system is largely due to the lack of data. However, it takes a lot of manpower and material resources to get labeled data, and using a small number of ``expensive'' labeled data instead of a large number of ``cheap'' unlabeled data is a huge waste of data resources. Considering the reasons mentioned above, we constructively propose adding pseudo ratings with high confidence improves the recommender robustness. 

Unfortunately, in the implicit recommendation system concerned in this paper, it is challenging to obtain high confidence pseudo scores. This is because the output of recommender systems is prediction scores, not confidence, unlike other areas of machine learning(e.g., in the image field, the output is the prediction probability). So we develop TCD, which uses three models and takes the prediction consistency ratings of any two models as the high confidence pseudo ratings of the remaining model. Moreover, the use of three models can not only provide confidence scores but also improve the robustness of the model through the collaborative training of three models. The framework is shown in Fig.1. In theory, more models with majority votes are more beneficial to obtain high-confidence data. However, the training of the model is linearly positively related to the number of models. We found that the performance of the three models is satisfactory and the training delay is tolerable. Now we provide details of the proposed TCD
for defending against poisoning attacks.

Let $D$ denote the dataset, $D_{L}$ denotes the scoring samples of $D$, where each sample $(u,i,r_{i,j})$ denotes that the user $u$'s rating on item $i$ is $r_{i,j}$, and $D_{U}$ denotes the no scoring samples of $D$, where each sample is like $(u,i)$. The goal of the recommendation system $h$ is to predict accurate scores $\hat{r}_{u,i}=h(u,i)$ of each sample $(u,i)\in D_U$.

In TCD, we denote the three models as $h_0$, $h_1$, and $h_2$, respectively. For any model,
if the predicted scores of the other two models are consistent, then we have reason to believe that the predicted scores are high-confident and reliable to be added to the training set which addresses the difficulty to measure rating confidence. For instance, if $h_{0}$ and $h_{1}$ agree on the
labeling $r_{i,j}$ of $(u,i)$ in $D_{U}$, then $(u,i,r_{i,j})$ will be put into the training set for $h_{2}$ as auxiliary training data. It
is obvious that in such a scheme if the prediction of $h_{0}$ and $h_{1}$
on $(u,i)$ is correct, then $h_{2}$ will receive a new sample with high confidence for
further training. This strategy takes into account that it is difficult for attackers to learn the real rating distribution, causing the poisoning profiles to deviate from the real data \cite{lin2020attacking}, which is reflected in the instability of their training. Therefore, cooperative training will magnify the influence of real profiles and relatively weaken the harm of false profiles.  

Besides, the predicted ratings are floating points, making it impractical to judge based on the consistent rating. So we define a projection function $\Pi(\cdot)$ to project continuous scores onto reasonable discrete scores. In this way, only when two models give the same rating on $(u,i)$ after projection, do we take the rating as the pseudo label and put $(u, i, \Pi(\hat{h_{j}}(u,i)))$ into the training set $D_{L}^{(k)}$.

\begin{algorithm}
	\caption{Triple Cooperative Defense}
	\LinesNumbered 
	\KwIn{The epochs of training $T$, the epochs of pre-training $T_{pre}$, three models $h_{1}(u,i), h_{2}(u,i), h_{3}(u,i)$, labeled data $D_{L}$, unlabeled data $D_{U}$, projection function $\Pi(x)$}
	\For{$T_{pre}$ epochs}{
		\For{$j \in [0,1,2]$}{
			Train $h_{j} $ based on the training set $D_{L}$
		}
	}
	
	\For{$T-T_{pre}$ epochs}{
		\For{$j \in [0,1,2]$}{
			$D_{L}^{(j)} \gets D_{L}$\\
			\For{every $(u,i) \in D_{U}$}{
				\If{$\Pi(\hat{h}_{(j+1)mod 3}(u,i)) = \Pi(\hat{h}_{(j+2)mod 3}(u,i))$}
				{$D_{L}^{(j)} \gets D_{L}^{(j)} \cup \{(u, i, \Pi(\hat{h}_{(j+1)mod 3}(u,i)))\}$\\
				}
			}
			Train $h_{j} $ based on training set $D_{L}^{(j)}$
		}
	}
	
\end{algorithm}

The algorithm of TCD is shown in Alg. 1. Each model is pre-trained from lines 1 to 5. Then, for each round of training for each model, an unlabeled prediction will be labeled if any two models agree on the labeling, as shown in lines 6 through 10. These pseudo labels with high confidence will be put into the third model's training dataset to reduce the harm that poisoning data do to the model, as shown in lines 11 through 16. After the training, we can perform the recommendation task using any model. Since the structure of each model is unchanged, the proposed strategy does not have inference delay, which is of more concern to practical applications.  

It is worth noting that in the pre-training phase, we used the same dataset $D_{L}$ for all models. Theoretically, we need to choose different training subsets to ensure the diversity of the model. This is necessary for other domains, such as computer version, because the number of parameters in a classifier is independent of the number of samples. However, in the recommender systems with extremely sparse data, selecting a subset means that a large number of users are cold-start users, and the parameters of these users cannot be trained, which directly leads to unsatisfactory recommendation performance. Therefore, all label data are selected for pre-training, while the models‘ diversity is guaranteed by different pseudo-labels in collaborative training. 
\section{EXPERIMENT}

\subsection{Settings}
\subsubsection{Datasets}
We use three real-world datasets commonly used
in the security studies \cite{christakopoulou2019adversarial,yuan2019adversarial} of the recommender system, including FilmTrust\footnote{https://www.librec.net/datasets/flmtrust.zip}, ML-100K\footnote{https://grouplens.org/datasets/movielens}
(MovieLens-100K), and ML-1M\footnote{https://grouplens.org/datasets/movielens}(MovieLens-1M). ML-100K includes 943 users who
have rated 1,682 movies for 100,000 ratings. ML-1M comprises 6,040
users who have rated 3,706 movies about one million times. For
FilmTrust, the same pretreatment as \cite{lin2020attacking} is used to filter cold-start
users who seriously affect the recommender system (the rating
number is less than 15), leaving 796 users with trust ratings for
2011 movies. Table 1 lists the detailed statistics of these datasets. All
ratings are from 1 to 5, and we normalized them to [0, 1] in the
experiments. For each dataset, we randomly select a positive sample
from each user for testing, and the rest are used as the training set
and verification set in a 9:1 ratio.

\begin{table}[htbp]
	\centering
	\caption{Statistics of datasets}
	\setlength{\tabcolsep}{5mm}{
	\begin{tabular}{ccccc}
		\toprule  
		Dataset&users&items&ratings&sparsity \\ 
		\midrule  
		FilmTrust&796&2011&30880&98.07\\
		ML-100K&943&1682&100000&93.70\\
		ML-1M&6040&3706&1000209&95.53\\
		\bottomrule  
	\end{tabular}}
\end{table}

\subsubsection{Attack Methods}
In the low-knowledge attacks studied in this paper, the attacker uses captured partial data to rebuild a local simulator which is similar to the target model. Then, the attacker take the local simulator as a white box for attacking. The validity of this setting is guaranteed by the transferability of the attack. Here we use the following attacks for robustness validation:
\begin{itemize}
	\item \textbf{Random Attack} \cite{lam2004shilling}: This attack assigns the maximum rating
	to the target item and rates selected items according to the normal
	distribution of all user ratings at random.
	
	\item \textbf{Average Attack} \cite{lam2004shilling}: The only difference from Random Attack
	is that the non-target selected item is randomly rated with the
	normal rating distribution of items.
	
	\item \textbf{AUSH Attack} \cite{lin2020attacking}: This attack uses GAN to generate fake
	users to carry out attacks imperceptibly and assigns the highest
	rating to the target item.
	
	\item \textbf{PGA Attack} \cite{li2016data}: This attack builds an attack objective and uses
	SGD to update the poisoned user’s ratings to optimize the objective.
	Finally, the first items with the largest ratings are selected as
	the fake user’s filler items.
	
	\item \textbf{TNA Attack} \cite{fang2020influence}: This attack selects a subset of the most influential users in the dataset and optimizes the rating gap between the
	target item and top-K items in the user subset. Here we use S-TNA. 
\end{itemize}

\subsubsection{Baselines}
We compare the proposed TCD with the following
robust algorithms:
\begin{itemize}
	\item \textbf{Adversarial Training(AT)}\cite{he2017neural}: In each training step, it first uses SGD to optimize the inner objective to generate small perturbations, adds them to the parameters, and then performs training.
	
	\item \textbf{Random Adversarial Training(RAT)}\cite{he2017neural}: In each training step, it first uses the truncated normal distribution N(0,0.01) to generate small perturbations, adds them to the parameters, and then performs training.
	
\end{itemize}
These methods cannot enjoy both generalization and robustness. The larger the noise is, the better the robustness will be, but the generalization will decrease significantly. Therefore, 0.03 is selected as a compromise.
\subsubsection{Evaluation Metric}
We first use HR@50 (Hit Ratio), just like \cite{wu2021fight}, which
calculates the proportion of test items that appear in the user’s top-50 recommendation list. Setting a large K helps make apparent comparisons between
defense methods and collaborative filtering is often used for candidate selection in practical recommendations, so it is more instructive to select a larger K to ensure a high recall \cite{he2018adversarial}. Besides, we use robustness improvement
$RI = 1- (HR_{defense} - HR_{orgin})/(HR_{attack} - HR_{orgin})$ defined in \cite{wu2021fight}. The closer
the value is to 1, the better the robustness. We report the average
results of 30 independent repeated experiments and perform paired
t-test to judge the statistical significance when necessary.

\subsubsection{Parameters Setting}
We concern with the MF-based collaborative filtering method, and we set the latent factor
dimension $d$ to 128, the batch size to 2048, and the regularization parameter to 0.005. In FilmTrust, ML-100K, and ML-1M, $Tpre$ is set to 1, 4, 2, respectively. The model is trained for 40 epochs, the results are based on the choice of the smallest MSE, and the
Adam optimizer is used for training. Besides, we set the attacker knowledge-cost to 0.4, the attack size to 3\%, and the pseudo-label rate of ML-1M to 0.2. For the target items of
attacks, we learn two types of items: (1) random items randomly
selected from all items, and (2) unpopular items randomly selected
from items with the number of rates less than 5. In each attack,
we set the number of target items to 5 and set the number of filler items $m                                                                                                                                                                                                                                                                                                                                                                                                                                                                                                        '$ to the average number of ratings per user. The source
code of TCD is available at https://github.com/greensun0830/TCD.

\begin{table}[htbp]
	\centering
	\caption{Attack performance under different attack knowledge-cost.}
	\begin{adjustbox}{width=1\textwidth}
		\small
		\begin{tabular}{c|c|c|c|c|c|c|c|c|c|c|c|c|c}
			\noalign{\smallskip} \hline \noalign{\smallskip}
			\multirow{3}[3]{*}{Dataset} & \multicolumn{7}{c|}{Random Items}                      & \multicolumn{6}{c}{Unpopular Items} \\
			\noalign{\smallskip} \cline{2-14} \noalign{\smallskip}
			& \multirow{2}[2]{*}{Attack} &\multirow{2}[2]{*}{Origin} & \multicolumn{5}{c|}{Attack Knowledge-cost}            & \multirow{2}[2]{*}{Origin} & \multicolumn{5}{c}{Attack Knowledge-cost} \\
			\noalign{\smallskip} \cline{4-8} \cline{10-14}\noalign{\smallskip} 
			&       &       & 0.2   & 0.4   & 0.6   & 0.8   & 1     &       & 0.2   & 0.4   & 0.6   & 0.8   & 1 \\
			\noalign{\smallskip} \hline \noalign{\smallskip}
			\multirow{5}[0]{*}{Filmtrust} & Average & 0.1617 & 0.0889 & 0.1005 & 0.1612 & 0.1222 & 0.1303 & 0.0000 & 0.0016 & 0.0022 & 0.0013 & 0.0017 & 0.0024 \\
			& Random & 0.1702 & 0.1376 & 0.1213 & 0.1622 & 0.1214 & 0.1404 & 0.0000 & 0.0034 & 0.0019 & 0.0012 & 0.0028 & 0.0023 \\
			& AUSH  & 0.1629 & 0.1132 & 0.1403 & 0.1625 & 0.2540 & 0.2675 & 0.0000 & 0.0152 & 0.0296 & 0.0285 & 0.0283 & 0.0461 \\
			& PGA   & 0.1574 & 0.0983 & 0.1031 & 0.1625 & 0.1471 & 0.1396 & 0.0000 & 0.0028 & 0.0013 & 0.0040 & 0.0049 & 0.0080 \\
			& TNA   & 0.1628 & 0.5619 & 0.5463 & 0.1446 & 0.5435 & 0.3380 & 0.0000 & 0.3054 & 0.4059 & 0.1839 & 0.0899 & 0.0807 \\
			\noalign{\smallskip} \hline \noalign{\smallskip}
			\multirow{5}[0]{*}{ML-100K} & Average & 0.0233 & 0.1829 & 0.1579 & 0.2193 & 0.2237 & 0.2209 & 0.0000 & 0.0255 & 0.1572 & 0.5094 & 0.5943 & 0.4694 \\
			& Random & 0.0234 & 0.0519 & 0.0956 & 0.0812 & 0.1099 & 0.0870 & 0.0000 & 0.1101 & 0.1056 & 0.1186 & 0.0906 & 0.0874 \\
			& AUSH  & 0.0233 & 0.1676 & 0.2819 & 0.3112 & 0.3667 & 0.3013 & 0.0000 & 0.0756 & 0.2320 & 0.7809 & 0.7942 & 0.8150 \\
			& PGA   & 0.0237 & 0.0855 & 0.1583 & 0.1673 & 0.1194 & 0.1667 & 0.0000 & 0.4558 & 0.2828 & 0.3912 & 0.3113 & 0.2809 \\
			& TNA   & 0.0244 & 0.0735 & 0.2355 & 0.2786 & 0.2512 & 0.2714 & 0.0000 & 0.6925 & 0.3934 & 0.6932 & 0.5628 & 0.7511 \\
			\noalign{\smallskip} \hline \noalign{\smallskip}
			\multirow{5}[0]{*}{ML-1M} & Average & 0.0000 & 0.1829 & 0.2390 & 0.2812 & 0.2674 & 0.3116 & 0.0000 & 0.9029 & 0.9326 & 0.9261 & 0.9408 & 0.9434 \\
			& Random & 0.0000 & 0.0519 & 0.0568 & 0.0563 & 0.0596 & 0.0608 & 0.0000 & 0.7213 & 0.7184 & 0.6471 & 0.7014 & 0.7588 \\
			& AUSH  & 0.0000 & 0.1676 & 0.2829 & 0.3145 & 0.3061 & 0.3278 & 0.0000 & 0.9680 & 0.9712 & 0.9767 & 0.9759 & 0.9803 \\
			& PGA   & 0.0000 & 0.0855 & 0.1027 & 0.1036 & 0.0418 & 0.0336 & 0.0000 & 0.9569 & 0.9433 & 0.9243 & 0.9034 & 0.9118 \\
			& TNA   & 0.0000 & 0.0735 & 0.2622 & 0.3046 & 0.3114 & 0.3406 & 0.0000 & 0.9068 & 0.9325 & 0.9395 & 0.9508 & 0.9496 \\
			\noalign{\smallskip} \hline \noalign{\smallskip}
		\end{tabular}%
	\end{adjustbox}
	\label{tab:addlabel}%
\end{table}%

\subsection{Result Analysis}
In this section, we compare the robustness and generalization of
the model configured with TCD and other defense methods.

\subsubsection{Attack Threat}
Different attack knowledge-cost leads to different attack performances, as shown in Table 2. We can find that a larger attack knowledge-cost does not have better attack performance, even when attackers only know 20\% of the model knowledge, they can achieve a good attack effect, and in most cases, 40\% attack knowledge performs well. Moreover, considering practical application scenarios, attacks cannot get full knowledge about recommender systems. So we choose to set the attack knowledge-cost to 0.4 to ensure its practicability while achieving a good attack performance. However, we also found that not all attacks are effective. For example, heuristic Random Attack and Average Attack are ineffective in FilmTrust and even reduce the exposure rate of target items, which emphasizes the significance of studying optimization-based attacks.  

\subsubsection{Robustness}

We evaluate the hit ratio of target items in attack
and defense, as shown in Table 3. The Origin denotes the unperturbed model, and the Attack represents the perturbed model with
no defense. Consistent with the findings in Table 3, 
We have the following finds: 

\begin{table*}[htbp]
	\centering
	\caption{The performance in target items (robustness). *, ** and *** indicate that the improvements over the best results of baselines are statistically significant for $ p<0.05, p<0.01 $ and $ p<0.001 $, respectively.}
	\begin{adjustbox}{width=1\textwidth}
		\small
		\begin{tabular}{c|c|c|c|c|c|c|c|c|c|c|c}
			\noalign{\smallskip} \hline \noalign{\smallskip}
			\multirow{2}[2]{*}{Dataset} & \multirow{2}[2]{*}{Attack} & \multicolumn{5}{c|}{Random Items}      & \multicolumn{5}{c}{Unpopular Items} \\
			\noalign{\smallskip} \cline{3-12} \noalign{\smallskip}
			&       & Origin & Attack & AT & RAT & TCD & Origin & Attack & AT & RAT & TCD \\
			\noalign{\smallskip} \hline \noalign{\smallskip}
			\multirow{5}[0]{*}{FilmTrust\newline{}} & Average & 0.1617 & 0.1005 & 0.0961 & 0.1001 &  \textbf{0.1093**} & 0.0000 & 0.0016 & 0.0010 & 0.0008 & \textbf{0.0009***} \\
			& Random & 0.1702 & 0.1213 & 0.1187 & 0.1257 & 0.1123 & 0.0000 & 0.0016 & 0.0020 & 0.0019 & \textbf{0.0016***} \\
			& AUSH  & 0.1629 & 0.1403 & 0.1345 & 0.1454 & 0.2204 & 0.0000 & 0.0323 & 0.0284 & 0.0330 & \textbf{0.0024***} \\
			& PGA   & 0.1574 & 0.1031 & 0.1008 & 0.1075 &  \textbf{0.1101**} & 0.0000 & 0.0012 & 0.0015 & 0.0016 & \textbf{0.0009***} \\
			& TNA   & 0.1628 & 0.5463 & 0.5346 & 0.5489 & \textbf{0.4086***} & 0.0000 & 0.4276 & 0.4251 & 0.4444 & \textbf{0.0416***} \\
			\noalign{\smallskip} \hline \noalign{\smallskip}
			\multirow{5}[0]{*}{ML-100K\newline{}} & Average & 0.0233 & 0.1579 & 0.1741 & 0.1525 & \textbf{0.0340***} & 0.0000 & 0.1694 & 0.1796 & 0.1421 & \textbf{0.0009***} \\
			& Random & 0.0234 & 0.0956 & 0.0932 & 0.0889 & \textbf{0.0353***} & 0.0000 & 0.1105 & 0.1277 & 0.0965 & \textbf{0.0010***} \\
			& AUSH  & 0.0233 & 0.2819 & 0.2665 & 0.2773 & \textbf{0.0355***} & 0.0000 & 0.1478 & 0.1869 & 0.1714 & \textbf{0.0009***} \\
			& PGA   & 0.0237 & 0.1583 & 0.1654 & 0.1471 & \textbf{0.0385***} & 0.0000 & 0.3486 & 0.4656 & 0.4108 & \textbf{0.0015***} \\
			& TNA   & 0.0244 & 0.2355 & 0.2411 & 0.2369 & \textbf{0.0334***} & 0.0000 & 0.3011 & 0.3624 & 0.2986 & \textbf{0.0015***} \\
			\noalign{\smallskip} \hline \noalign{\smallskip}
			\multirow{5}[0]{*}{ML-1M} & Average & 0.0000 & 0.2390 & 0.1547 & 0.2364 & \textbf{0.0048***} & 0.0000 & 0.8604 & 0.8593 & 0.8633 & \textbf{0.0309***} \\
			& Random & 0.0000 & 0.0568 & 0.0455 & 0.0517 & \textbf{0.0072***} & 0.0000 & 0.6513 & 0.6450 & 0.6064 & \textbf{0.0254***} \\
			& AUSH  & 0.0000 & 0.2829 & 0.1395 & 0.2518 & \textbf{0.0031***} & 0.0000 & 0.9056 & 0.8845 & 0.8999 &\textbf{0.0353***} \\
			& PGA   & 0.0000 & 0.1027 & 0.0832 & 0.0968 & \textbf{0.0191***} & 0.0000 & 0.8577 & 0.8501 & 0.8498 & \textbf{0.0236***} \\
			& TNA   & 0.0000 & 0.2622 & 0.1663 & 0.2403 & \textbf{0.0042***} & 0.0000 & 0.8654 & 0.8525 & 0.8650 &\textbf{0.0269***} \\
			\noalign{\smallskip} \hline \noalign{\smallskip}
		\end{tabular}%
	\end{adjustbox}
\end{table*}%

\begin{figure*}[h]
	\centering
	\subfloat[FilmTrust]{\includegraphics[width=0.5\linewidth]{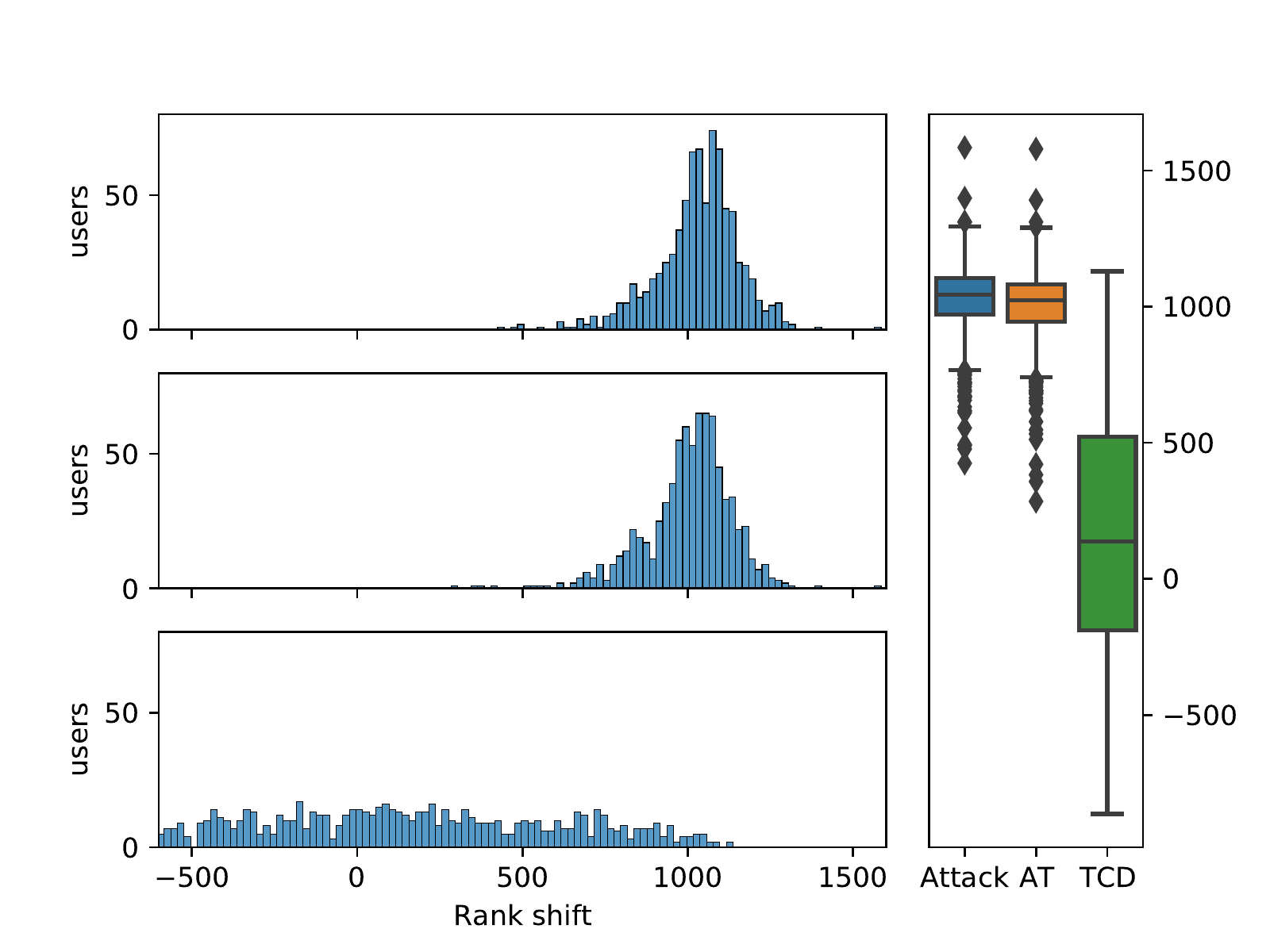}}
	\hfill
	\subfloat[ML-100K]{\includegraphics[width=0.5\linewidth]{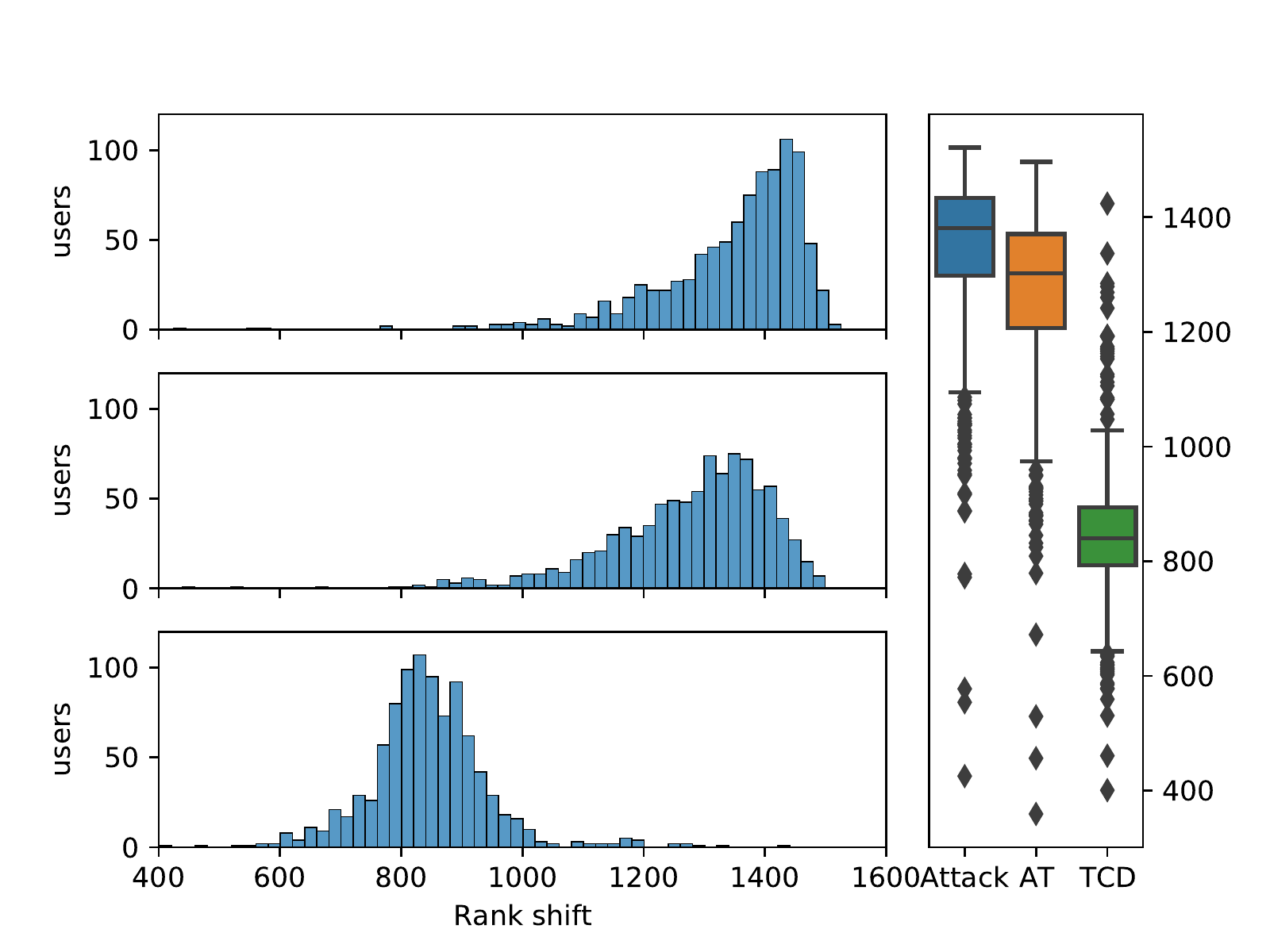}}
	\hfill
	\caption{The distribution of rank shift. In FilmTrust and Ml-100k, top: TNA attack; middle: AT on TNA attack; bottom:
		TCD on TNA attack; boxplot: statistical distributions of rank shift. The closer the rank shift is to 0,
		the smaller the damage of the attack}
	\label{fig:ri}
\end{figure*}

\begin{itemize}
	\item These defense methods are positive in weakening
	the attack’s damage concerning HR in most cases. 
	\item The
	proposed TCD achieves remarkable defense results, almost close
	to the unperturbed model performance. On average, we reduce
	the impact of attacks on random items by over 88\% and unpopular
	items by over 82\%, which effortlessly outperforms baselines. 
	\item We notice that the performance of TCD against Average and Random on FilmTrust's unpopular items is slightly
	inferior when compared with the defense against other attacks while almost every performance of TCD on ML-100k and ML-1M is better than that of baselines. We
	suspect that Filmtrust is too small to represent real data, making it easier for adversarial training to
	discover and learn adversary data non-robust features while making it more formidable for TCD to find data’s non-robust features.
\end{itemize}
Besides, Fig. 2 shows the Rank shift distribution of target items (unpopular items) under the TNA attack. The attack significantly promotes the target item’s
rank among all users. After using adversarial training, the rank change caused by the attack can be eased, but it is only slight. On
the contrary, TCD impels the distribution of rank shift obviously
tends to 0, which means that applying TCD can produce more stable
recommendations in a disturbed environment.
In conclusion, these results confirm the positive effect of TCD in
boosting recommendation robustness against poisoning attacks.

\subsubsection{Generalization}

\begin{table*}[htbp]
	\centering
	\caption{The performance in test set (generalization). *, ** and *** indicate that the improvements over the unperturbed model are statistically significant for $ p<0.05, p<0.01 $ and $ p<0.001 $, respectively.}
	\begin{adjustbox}{width=1\textwidth}
		\small
		\begin{tabular}{c|c|c|c|c|c|c|c|c|c|c|c}
			\noalign{\smallskip} \hline \noalign{\smallskip}
			\multirow{2}[2]{*}{Dataset} & \multirow{2}[2]{*}{Attack} & \multicolumn{5}{c|}{Random Items}      & \multicolumn{5}{c}{Unpopular Items} \\
			\noalign{\smallskip} \cline{3-12} \noalign{\smallskip}         &       & Origin & Attack & AT & RAT & TCD & Origin & Attack & AT & RAT & TCD \\
			\noalign{\smallskip} \hline \noalign{\smallskip}
			\multirow{5}[0]{*}{Filmtrust}& Average & 0.8253 & 0.8196 & 0.8086 & 0.8187 & \textbf{0.8640***} & 0.8273 & 0.8258 & 0.8096 & 0.8160 & \textbf{0.8648***} \\
			& Random & 0.8266 & 0.8245 & 0.8010 & 0.8179 & \textbf{0.8635***} & 0.8275 & 0.8221 & 0.8085 & 0.8185 & \textbf{0.8651***} \\
			& AUSH  & 0.8252 & 0.8240 & 0.8170 & 0.8193 & \textbf{0.8660***} & 0.8256 & 0.8196 & 0.8053 & 0.8184 & \textbf{0.8639***} \\
			& PGA   & 0.8257 & 0.8222 & 0.8046 & 0.8181 & \textbf{0.8643***} & 0.8266 & 0.8212 & 0.8088 & 0.8205 & \textbf{0.8636***} \\
			& TNA   & 0.8264 & 0.8079 & 0.7840 & 0.8021 & \textbf{0.8639***} & 0.8273 & 0.8054 & 0.7824 & 0.8002 &\textbf{0.8622***} \\
			\noalign{\smallskip} \hline \noalign{\smallskip}
			\multirow{5}[0]{*}{ML-100K} & Average & 0.2006 & 0.1985 & 0.1907 & 0.1995 & \textbf{0.2875***} & 0.2020 & 0.1998 & 0.1924 & 0.1970 &\textbf{0.2836***} \\
			& Random & 0.1988 & 0.2025 & 0.2003 & 0.2005 & \textbf{0.2898***} & 0.1969 & 0.2058 & 0.1973 & 0.2003 & \textbf{0.2804***} \\
			& AUSH  & 0.1998 & 0.1978 & 0.1894 & 0.1940 & \textbf{0.2824***} & 0.2007 & 0.1971 & 0.1904 & 0.1951 & \textbf{0.2797***} \\
			& PGA   & 0.2005 & 0.1960 & 0.1872 & 0.1920 & \textbf{0.2851***} & 0.2022 & 0.1887 & 0.1848 & 0.1905 & \textbf{0.2858***} \\
			& TNA   & 0.1995 & 0.1990 & 0.1933 & 0.1967 & \textbf{0.2897***} & 0.2003 & 0.1970 & 0.1873 & 0.1911 & \textbf{0.2785***} \\
			\noalign{\smallskip} \hline \noalign{\smallskip}
			\multirow{5}[0]{*}{ML-1M} & Average & 0.0834 & 0.0748 & 0.0542 & 0.0718 & \textbf{0.1097***} & 0.0843 & 0.0718 & 0.0519 & 0.0695 & \textbf{0.1094***} \\
			& Random & 0.0844 & 0.0877 & 0.0850 & 0.0860 & \textbf{0.1097***} & 0.0833 & 0.0827 & 0.0816 & 0.0830 & \textbf{0.1097***} \\
			& AUSH  & 0.0837 & 0.0733 & 0.0523 & 0.0689 & \textbf{0.1105***} & 0.0832 & 0.0685 & 0.0448 & 0.0639 & \textbf{0.1103***} \\
			& PGA   & 0.0837 & 0.0842 & 0.0805 & 0.0818 & \textbf{0.1092***} & 0.0831 & 0.0774 & 0.0731 & 0.0767 & \textbf{0.1105***} \\
			& TNA   & 0.0827 & 0.0748 & 0.0540 & 0.0713 & \textbf{0.1101***} & 0.0846 & 0.0743 & 0.0539 & 0.0719 & \textbf{0.1096***} \\
			\noalign{\smallskip} \hline \noalign{\smallskip}
		\end{tabular}%
	\end{adjustbox}
	\label{tab:ex:cref}
\end{table*}%

It is meaningless to improve the robustness
at the cost of apparently sacrificing the generalization of standard
recommendations. Table 4 records the HR of various defense methods in the holdout test set. We have the following finds: 
\begin{itemize}
	\item TCD surprisingly improves the generalization of the three datasets and the improvement is above 0.02 in terms of HR.
	
	These results confirm that TCD effectively guarantees the model’s
	generalization while performing high-quality defense. 
\end{itemize}

\subsection{Parameter Analysis}

\subsubsection{Performance under Different Attack knowledge-cost}
We conduct the robustness improvement test of TCD under different
attack knowledge-cost, as illustrated in Fig. 3. On the one hand, the overall
defense performance of TCD remains at a high level, although there
will be individual cases on FilmTrust where it performs not that well. On the other hand, as the
attack intensity increases, the robustness against attacks is still satisfactory. Especially in ML-100K and ML-1M, RI is almost clear 100\%!

\begin{figure}[h]
	\centering
	\includegraphics[width=1.\columnwidth]{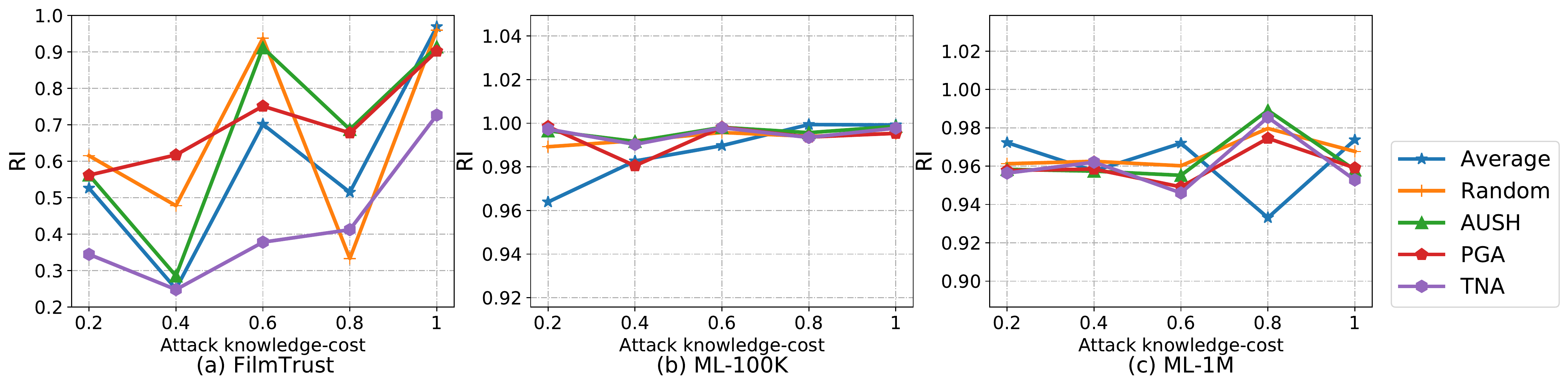}
	\caption{Robustness improvement under different attack knowledge-cost
	}
	\label{fig:knowledge cost}
\end{figure}

\subsubsection{Performance under Different Pseudo-label Ratios}
The training time of TCD is directly proportional to the training set. Considering the size and sparsity of ML-1M, we decide to put only part of pseudo labels into the training set, and we denote the pseudo labels rate as the proportion of the pseudo labels which is put into the training set. We conduct the robustness test of TCD under different
pseudo-label ratios, as illustrated in Fig. 4. With the injected pseudo-label ratio increases, the robustness of the model is improved accordingly, and 0 on the abscissa means attack without any defense. When the pseudo-label ratio is only 0.2, TCD can significantly improve the robustness of the model, which emphasizes its practicality in large datasets.  

\begin{figure}[h]
	\centering
	\includegraphics[width=0.9\columnwidth]{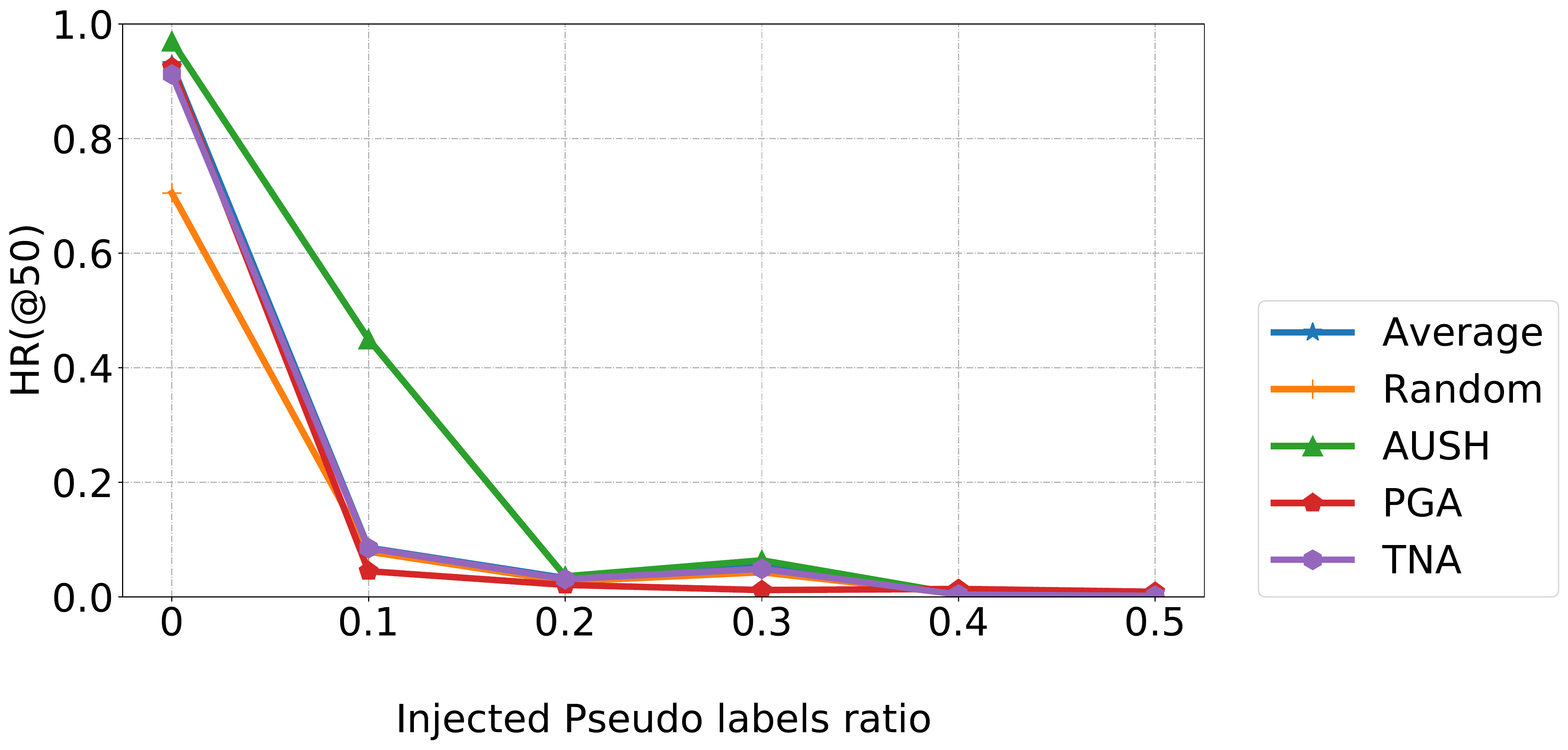}
	\caption{The defense performance of the target items on ML-1M under different injected pseudo-label ratio}
	\label{fig:pseudo labels rate}
\end{figure}

\section{Conclusion and Outlook}
In this paper, we proposed the TCD method to defend against attacks on recommender systems. It is noteworthy that TCD is the first algorithm to combine data-processing-based defense with model-based defense in recommender systems. Specifically, we sequentially use the high-confidence prediction ratings of any two models as auxiliary training data for the remaining models. Since TCD enhances data by adding pseudo labels instead of deleting abnormal data, it can avoid cleaning normal data and train a more accurate and robust model. Moreover, the cooperative training of the three models makes it beneficial for model generalization. Moreover, TCD is a general framework, so it can be combined with other defense methods. In the future, we plan to apply TCD in non-recommendation fields.

\renewcommand{\bibsection}{\section*{References}} 
\bibliographystyle{splncsnat}
\begingroup
  \microtypecontext{expansion=sloppy}
  \small 
  \bibliography{paper}

\begin{thebibliography}{40}
\providecommand{\natexlab}[1]{#1}
\providecommand{\url}[1]{\texttt{#1}}
\providecommand{\urlprefix}{}

\bibitem[{Athalye et~al.(2018)Athalye, Carlini, and
  Wagner}]{athalye2018obfuscated}
Athalye, A., Carlini, N., Wagner, D.: Obfuscated gradients give a false sense
  of security: Circumventing defenses to adversarial examples.
\newblock In: ICML. pp. 274--283. PMLR (2018)

\bibitem[{Bobadilla et~al.(2013)Bobadilla, Ortega, Hernando, and
  Guti{\'e}rrez}]{bobadilla2013recommender}
Bobadilla, J., Ortega, F., Hernando, A., Guti{\'e}rrez, A.: Recommender systems
  survey.
\newblock Knowledge-based systems 46, 109--132 (2013)

\bibitem[{Burke et~al.(2006)Burke, Mobasher, Williams, and
  Bhaumik}]{burke2006classification}
Burke, R., Mobasher, B., Williams, C., Bhaumik, R.: Classification features for
  attack detection in collaborative recommender systems.
\newblock In: KDD. pp. 542--547 (2006)

\bibitem[{Cao et~al.(2013)Cao, Wu, Mao, and Zhang}]{cao2013shilling}
Cao, J., Wu, Z., Mao, B., Zhang, Y.: Shilling attack detection utilizing
  semi-supervised learning method for collaborative recommender system.
\newblock WWW 16(5-6), 729--748 (2013)

\bibitem[{Chevalier and Mayzlin(2006)}]{chevalier2006effect}
Chevalier, J.A., Mayzlin, D.: The effect of word of mouth on sales: Online book
  reviews.
\newblock J Mark Res 43(3), 345--354 (2006)

\bibitem[{Christakopoulou and Banerjee(2019)}]{christakopoulou2019adversarial}
Christakopoulou, K., Banerjee, A.: Adversarial attacks on an oblivious
  recommender.
\newblock In: RecSys. pp. 322--330 (2019)

\bibitem[{Deldjoo et~al.(2021)Deldjoo, Noia, and Merra}]{deldjoo2021survey}
Deldjoo, Y., Noia, T.D., Merra, F.A.: A survey on adversarial recommender
  systems: from attack/defense strategies to generative adversarial networks.
\newblock CSUR 54(2), 1--38 (2021)

\bibitem[{Du et~al.(2018)Du, Fang, Yi, Xu, Cheng, and Tao}]{du2018enhancing}
Du, Y., Fang, M., Yi, J., Xu, C., Cheng, J., Tao, D.: Enhancing the robustness
  of neural collaborative filtering systems under malicious attacks.
\newblock IEEE Trans. Multimedia 21(3), 555--565 (2018)

\bibitem[{Fan et~al.(2021)Fan, Derr, Zhao, Ma, Liu, Wang, Tang, and
  Li}]{fan2021attacking}
Fan, W., Derr, T., Zhao, X., Ma, Y., Liu, H., Wang, J., Tang, J., Li, Q.:
  Attacking black-box recommendations via copying cross-domain user profiles.
\newblock In: ICDE. pp. 1583--1594. IEEE (2021)

\bibitem[{Fang et~al.(2020)Fang, Gong, and Liu}]{fang2020influence}
Fang, M., Gong, N.Z., Liu, J.: Influence function based data poisoning attacks
  to top-n recommender systems.
\newblock In: Proceedings of The Web Conference 2020. pp. 3019--3025 (2020)

\bibitem[{Fang et~al.(2018)Fang, Yang, Gong, and Liu}]{fang2018poisoning}
Fang, M., Yang, G., Gong, N.Z., Liu, J.: Poisoning attacks to graph-based
  recommender systems.
\newblock In: ACSAC. pp. 381--392 (2018)

\bibitem[{Guo et~al.(2017)Guo, Tang, Ye, Li, and He}]{guo2017deepfm}
Guo, H., Tang, R., Ye, Y., Li, Z., He, X.: Deepfm: a factorization-machine
  based neural network for ctr prediction.
\newblock arXiv  (2017)

\bibitem[{He et~al.(2018)He, He, Du, and Chua}]{he2018adversarial}
He, X., He, Z., Du, X., Chua, T.S.: Adversarial personalized ranking for
  recommendation.
\newblock In: SIGIR. pp. 355--364 (2018)

\bibitem[{He et~al.(2017)He, Liao, Zhang, Nie, Hu, and Chua}]{he2017neural}
He, X., Liao, L., Zhang, H., Nie, L., Hu, X., Chua, T.S.: Neural collaborative
  filtering.
\newblock In: WWW. pp. 173--182 (2017)

\bibitem[{Himeur et~al.(2022)Himeur, Sayed, Alsalemi, Bensaali, Amira,
  Varlamis, Eirinaki, Sardianos, and Dimitrakopoulos}]{himeur2022blockchain}
Himeur, Y., Sayed, A., Alsalemi, A., Bensaali, F., Amira, A., Varlamis, I.,
  Eirinaki, M., Sardianos, C., Dimitrakopoulos, G.: Blockchain-based
  recommender systems: Applications, challenges and future opportunities.
\newblock Computer Science Review 43, 100439 (2022)

\bibitem[{Jin et~al.(2020)Jin, Lian, Liu, Liu, Ma, Xie, and
  Chen}]{jin2020sampling}
Jin, B., Lian, D., Liu, Z., Liu, Q., Ma, J., Xie, X., Chen, E.:
  Sampling-decomposable generative adversarial recommender.
\newblock Adv Neur In 33, 22629--22639 (2020)

\bibitem[{Lam and Riedl(2004)}]{lam2004shilling}
Lam, S.K., Riedl, J.: Shilling recommender systems for fun and profit.
\newblock In: WWW. pp. 393--402 (2004)

\bibitem[{Li et~al.(2016)Li, Wang, Singh, and Vorobeychik}]{li2016data}
Li, B., Wang, Y., Singh, A., Vorobeychik, Y.: Data poisoning attacks on
  factorization-based collaborative filtering.
\newblock NIPS 29, 1885--1893 (2016)

\bibitem[{Li et~al.(2020)Li, Wu, and Wang}]{li2020adversarial}
Li, R., Wu, X., Wang, W.: Adversarial learning to compare: Self-attentive
  prospective customer recommendation in location based social networks.
\newblock In: WSDM. pp. 349--357 (2020)

\bibitem[{Lian et~al.(2020)Lian, Wu, Ge, Xie, and Chen}]{lian2020geography}
Lian, D., Wu, Y., Ge, Y., Xie, X., Chen, E.: Geography-aware sequential
  location recommendation.
\newblock In: Proceedings of KDD'20. pp. 2009--2019 (2020)

\bibitem[{Lin et~al.(2020)Lin, Chen, Li, Xiao, and Li}]{lin2020attacking}
Lin, C., Chen, S., Li, H., Xiao, Y., Li, Lianyun \textbf{}and~Yang, Q.:
  Attacking recommender systems with augmented user profiles.
\newblock In: CIKM. pp. 855--864 (2020)

\bibitem[{Liu et~al.(2014)Liu, Hu, Mian, Tian, and Zhu}]{liu2014new}
Liu, H., Hu, Z., Mian, A., Tian, H., Zhu, X.: A new user similarity model to
  improve the accuracy of collaborative filtering.
\newblock KBS 56, 156--166 (2014)

\bibitem[{Machado et~al.(2021)Machado, Silva, and
  Goldschmidt}]{machado2021adversarial}
Machado, G.R., Silva, E., Goldschmidt, R.R.: Adversarial machine learning in
  image classification: A survey toward the defender’s perspective.
\newblock CSUR (1), 1--38 (2021)

\bibitem[{Madry et~al.(2017)Madry, Makelov, Schmidt, Tsipras, and
  Vladu}]{madry2017towards}
Madry, A., Makelov, A., Schmidt, L., Tsipras, D., Vladu, A.: Towards deep
  learning models resistant to adversarial attacks.
\newblock arXiv  (2017)

\bibitem[{Mobasher et~al.(2007)Mobasher, Burke, Bhaumik, and
  Williams}]{mobasher2007toward}
Mobasher, B., Burke, R., Bhaumik, R., Williams, C.: Toward trustworthy
  recommender systems: An analysis of attack models and algorithm robustness.
\newblock TOIT 7(4), 23--es (2007)

\bibitem[{Oh and Kumar(2022)}]{oh2022robustness}
Oh, S., Kumar, S.: Robustness of deep recommendation systems to untargeted
  interaction perturbations.
\newblock arXiv  (2022)

\bibitem[{Ovaisi et~al.(2022)Ovaisi, Heinecke, Li, Zhang, Zheleva, and
  Xiong}]{ovaisi2022rgrecsys}
Ovaisi, Z., Heinecke, S., Li, J., Zhang, Y., Zheleva, E., Xiong, C.: Rgrecsys:
  A toolkit for robustness evaluation of recommender systems.
\newblock arXiv  (2022)

\bibitem[{Park and Chang(2019)}]{park2019adversarial}
Park, D.H., Chang, Y.: Adversarial sampling and training for semi-supervised
  information retrieval.
\newblock In: The World Wide Web Conference. pp. 1443--1453 (2019)

\bibitem[{Si and Li(2020)}]{si2020shilling}
Si, M., Li, Q.: Shilling attacks against collaborative recommender systems: a
  review.
\newblock Artif Intell Rev 53(1), 291--319 (2020)

\bibitem[{Song et~al.(2020)Song, Li, Hu, Wu, Li, Li, and
  Gao}]{song2020poisonrec}
Song, J., Li, Z., Hu, Z., Wu, Y., Li, Z., Li, J., Gao, J.: Poisonrec: an
  adaptive data poisoning framework for attacking black-box recommender
  systems.
\newblock In: ICDE. pp. 157--168. IEEE (2020)

\bibitem[{Tang et~al.(2020)Tang, Wen, and Wang}]{tang2020revisiting}
Tang, J., Wen, H., Wang, K.: Revisiting adversarially learned injection attacks
  against recommender systems.
\newblock In: RecSys. pp. 318--327 (2020)

\bibitem[{Tang et~al.(2019)Tang, Du, He, Yuan, Tian, and
  Chua}]{tang2019adversarial}
Tang, J., Du, X., He, X., Yuan, F., Tian, Q., Chua, T.S.: Adversarial training
  towards robust multimedia recommender system.
\newblock IEEE Transactions on Knowledge and Data Engineering 32(5), 855--867
  (2019)

\bibitem[{Wu et~al.(2021{\natexlab{a}})Wu, Lian, Ge, Zhu, and
  Chen}]{wu2021triple}
Wu, C., Lian, D., Ge, Y., Zhu, Z., Chen, E.: Triple adversarial learning for
  influence based poisoning attack in recommender systems.
\newblock In: Proceedings of KDD'21. pp. 1830--1840 (2021{\natexlab{a}})

\bibitem[{Wu et~al.(2021{\natexlab{b}})Wu, Lian, Ge, Zhu, Chen, and
  Yuan}]{wu2021fight}
Wu, C., Lian, D., Ge, Y., Zhu, Z., Chen, E., Yuan, S.: Fight fire with fire:
  Towards robust recommender systems via adversarial poisoning training.
\newblock In: SIGIR. pp. 1074--1083 (2021{\natexlab{b}})

\bibitem[{WU et~al.(2014)WU, WANG, and CAO}]{wu2014survey}
WU, Z., WANG, Y., CAO, J.: A survey on shilling attack models and detection
  techniques for recommender systems.
\newblock Chinese Sci Bull 59(7), 551--560 (2014)

\bibitem[{Yang et~al.(2017)Yang, Gong, and Cai}]{yang2017fake}
Yang, G., Gong, N.Z., Cai, Y.: Fake co-visitation injection attacks to
  recommender systems.
\newblock In: NDSS (2017)

\bibitem[{Yuan et~al.(2019)Yuan, Yao, and Benatallah}]{yuan2019adversarial}
Yuan, F., Yao, L., Benatallah, B.: Adversarial collaborative neural network for
  robust recommendation.
\newblock In: SIGIR. pp. 1065--1068 (2019)

\bibitem[{Zhang et~al.(2018)Zhang, Zhang, Zhang, and Wang}]{zhang2018ud}
Zhang, F., Zhang, Z., Zhang, P., Wang, S.: Ud-hmm: An unsupervised method for
  shilling attack detection based on hidden markov model and hierarchical
  clustering.
\newblock Knowledge-Based Systems 148, 146--166 (2018)

\bibitem[{Zhang et~al.(2020)Zhang, Xu, Han, Niu, Cui, Sugiyama, and
  Kankanhalli}]{zhang2020attacks}
Zhang, J., Xu, X., Han, B., Niu, G., Cui, L., Sugiyama, M., Kankanhalli, M.:
  Attacks which do not kill training make adversarial learning stronger.
\newblock In: ICML. pp. 11278--11287. PMLR (2020)

\bibitem[{Zhang and Kulkarni(2014)}]{zhang2014detection}
Zhang, Z., Kulkarni, S.R.: Detection of shilling attacks in recommender systems
  via spectral clustering.
\newblock In: FUSION. pp. 1--8. IEEE (2014)

\end{thebibliography}
\endgroup

\ \\
%



\end{document}